\documentclass{article}

\usepackage{PRIMEarxiv}

\usepackage[utf8]{inputenc} % allow utf-8 input
\usepackage[T1]{fontenc}    % use 8-bit T1 fonts
\usepackage{hyperref}       % hyperlinks
\usepackage{url}            % simple URL typesetting
\usepackage{booktabs}  
\usepackage{tabularx}% professional-quality tables
\usepackage{amsfonts}       % blackboard math symbols
\usepackage{nicefrac}
\usepackage{threeparttable}
\usepackage{pdflscape}   % or \usepackage{lscape}
\usepackage{longtable}
\usepackage{caption} % compact symbols for 1/2, etc.
\usepackage{microtype}      % microtypography
\usepackage{lipsum}
\usepackage{fancyhdr} 
\usepackage{soul}
\usepackage{xcolor}
\usepackage[dvipsnames]{xcolor}
\sethlcolor{Apricot} % for highlighting
\usepackage{cite} % header
\usepackage{graphicx}       % graphics
\graphicspath{{media/}}     % organize your images and other figures under media/ folder

\newcommand{\thl}[2]{%
  \colorbox{#1!30}{#2} % 30% tint of the color
}

\title{Advancing site-specific disease and pest management in precision agriculture:~From reasoning-based foundation models to adaptive, feedback-driven learning
}

\author{
  Nitin Rai \\
  Department of Horticultural Sciences \\
  Gulf Coast Research and Education Center \\ University of Florida \\
  Wimauma, FL, USA \\
  \texttt{\{nitin.rai\}ufl.edu} \\
  \And
  Daeun (Dana) Choi \\
  Department of Agricultural and Biological Engineering  \\
  Gulf Coast Research and Education Center \\ University of Florida \\
  Wimauma, FL, USA \\
  \And
  Nathan S. Boyd \\
  Department of Horticultural Sciences \\ Gulf Coast Research and Education Center \\
  University of Florida \\
  Wimauma, FL, USA \\
  \And
  Arnold W. Schumann \\
  Department of Soil, Water, and Ecosystem Sciences \\ Citrus Research and Education Center \\
  University of Florida \\
  Wimauma, FL, USA \\
}

\begin{document}
\maketitle

\begin{abstract}
Site-specific disease management (SSDM) in agricultural crops has witnessed tremendous advancements over the past few decades using conventional machine and deep learning (ML \& DL) approaches for real-time computer vision applications. This research has evolved from handcrafted feature extraction to large-scale automated feature learning in single-modality datasets. However, with the rise of foundation models (FMs), the way large-scale crop disease datasets are processed is being fundamentally transformed. Unlike traditional convolutional neural networks, which often struggle to integrate multi-modal data or connect visual and textual information, FMs enable deeper level of understanding. They can interpret disease symptoms described in text, reason about relationships between symptoms and management factors, and even support interactive Q\&A platforms for growers and extension educators. Likewise, the integration of adaptive and imitation learning in robotics is enabling novel applications for field-based disease management. These emerging shifts in advanced computer vision and robotics research are redefining how crops are monitored and managed in in-field settings. Therefore, this study reviewed $\approx$40 research articles (with appropriate screening criteria) to highlight the application of FMs for SSDM in crops, focusing on two primary themes: large-language models (LLMs), and vision language models (VLMs). Additionally, the extended role of FMs in enabling adaptive learning (AL), reinforcement learning (RL), and digital twin frameworks for robotics-based targeted spraying is also discussed. Based on the results and discussion, several key conclusions emerge from this review: (a) FMs are gaining traction, with a notable increase in reported technical literature during 2023-24, (b) researchers are leveraging VLMs more than LLMs, with a five-to tenfold increase in published articles from 2023 to 2024, (c) approaches such as RL or AL are still in their infancy for developing smart spraying technologies that can learn from experience, (d) the integration of digital twins with RL in cyber-physical systems could offer a transformative approach for simulating targeted spraying in virtual environments, (e) addressing the sim-to-real gap, the performance drop when models trained in simulated or controlled environments are deployed in real-world disease scenarios, will be critical for ensuring robust and scalable management systems, (f) while perception models for disease detection are advancing, human-robot collaboration in crop disease management remains limited, particularly in leveraging human-in-the-loop approaches where robots autonomously detect early symptoms and humans validate uncertain cases, and (g) continued advancements in FMs, along with multi-modal integration and real-time feedback, are expected to drive the next generation of SSDM technologies.~For updates, resources, and contributions, please visit our \href{https://github.com/nitin-dominic/AgriPathogenDatabase}{\textcolor{orange}{AgriPathogenDatabase}} GitHub repository and consider submitting papers, code, or datasets.

%\lipsum[1]
\end{abstract}

% keywords can be removed

\section*{\centering Abbreviations}
\vspace{-0.45cm}
\begin{table}[h]
\centering
\tiny
   \caption*{The table below includes a list of abbreviations and its subsequent descriptions used in the review study.}
   \renewcommand{\arraystretch}{1}
      \resizebox{\textwidth}{!}{%
            \begin{tabular}{|p{4cm}|>{\arraybackslash}p{7cm}|}
      \toprule
      \textbf{Abbreviations} & \textbf{Descriptions} \\ 
      \midrule
      AI & Artificial intelligence \\
      AL & Adaptive learning \\
      BERT & Bidirectional encoder representations from transformers \\
      CLIP & Contrastive language-image pre-training \\
      CNN & Convolutional neural network \\
      DINO & DIstillation with NO labels \\
      DL & Deep learning  \\
      DT & Digital twin \\
      FM & Foundation model \\
      GAN & Generative adversarial network \\ 
      GPT & Generative pre-trained transformer \\
      HS & Hyperspectral \\
      IL & Imitation learning \\
      IoT & Internet of things \\
      LLM & Large language models \\
      LoRA & Low-Rank Adaptation \\
      ML & Machine learning \\
      MS & Multispectral \\
      NLP & Natural language processing \\
      RGB & Red, Green, and Blue \\
      RL & Reinforcement learning \\
      SAM & Segment anything model \\
      SD & Stable diffusion \\
      SSDM & Site-specific disease management \\
      UAV & Unmanned aerial vehicle \\
      ViT & Vision transformer \\
      VLA & Vision-language-action \\
      VLM & Vision language models \\
      YOLO & You Only Look Once \\
      \bottomrule
      \end{tabular}
      }
      %\label{tab1}
% \end{threeparttable}
 \end{table}

\section{Introduction}

The application of artificial intelligence (AI) has transformed agricultural automation across diverse domains, including weed classification for spot spraying ~\cite{wu2025review,rai2023weed}, crop yield estimation~\cite{sun2020multilevel,xiong2024daily}, disease monitoring~\cite{nandhini2022deep,hu2025amf}, phenotyping~\cite{rui2024high,qi2025multiscale}, and plant breeding~\cite{uzal2018seed,zhang2024dsbean}.~Among these, AI-driven computer vision models have emerged as particularly impactful in crop disease and pest management, where accurate visual recognition is pivotal for early detection and targeted intervention. These vision-based systems are increasingly integrated with smart sprayers and autonomous robots, enabling site-specific fungicide application while reducing chemical inputs and operational costs~\cite{liu2024review,yang2025advanced}. A major driver of this precision capability lies in the synergy between computer vision models deployed on edge systems, which provides machines with the ability to perceive and interpret complex agricultural environments.

At the heart of this development is data, large-scale image datasets, combined with the representation power of deep learning (DL) models, have been pivotal in transmitting “vision” to agricultural robots, thereby bridging the gap between raw perception and actionable intelligence in the field. Despite these advancements that DL models, particularly convolutional neural networks (CNNs), have achieved to address site-specific disease management (SSDM) in crops, they face notable limitations when deployed in agricultural environments. Their performance is often constrained by the availability of annotated datasets, which are expensive and time-consuming to curate for the wide variety of crops, diseases, and field conditions. Moreover, CNN models often require fine-tuning or transfer learning approaches when applied to new crops, disease symptoms, or environmental settings. These models are also limited to applications where multi-modalities of data is generated, as they can only take one format of data at a time. Such constraints have highlighted the need for more adaptive and generalizable approaches, where foundation models (FMs), trained on massive multi-modal datasets demonstrate clear advantage.

The term ``foundation models'' was first used by researchers at Stanford University~\cite{bommasani2021opportunities}. FMs are pretrained on massive and diverse datasets, have emerged as a transformative paradigm in computer vision, with the potential to overcome many limitations posed by CNN models. Unlike conventional models, FMs such as Contrastive Language-Image Pre-training (CLIP), Segment Anything (SAM)~\cite{kirillov2023segment,pathak2023autonomous}, and Generative Pretrained Transformer (GPT)~\cite{achiam2023gpt} architectures, are inherently versatile as they can be adapted to downstream agricultural tasks with minimal retraining and are capable of few-shot and even zero-shot learning~\cite{parnami2022learning,yu2025learning}. Several studies demonstrate their utility in crop disease detection and synthetic data generation, showcasing improved generalization and robustness compared to traditional task-specific models. For instance, the PlantCaFo model was developed using a few-shot approach that leverages the model's prior knowledge~\cite{jiang2025plantcafo}. Another work was inspired by the CLIP architecture and used the Progressive Mixup Prompt Leaning (PMPL) framework, which integrates hierarchical feature Mixup with prompt learning for crop disease recognition~\cite{chen2025enhancing}. Their ability to encode broad semantic understanding with the help of text-based prompts has introduced a step-change in how agricultural tasks can be analyzed, enabling models to move beyond narrow classification tasks toward more context-aware decision support tools. With the help of FMs, users can now access interactive Q\&A interfaces where they upload an image of a crop showing disease symptoms, and the model can reason about the presence of a particular disease, unlike conventional models trained only on specific labels\cite{quoc2025vision,lan2023visual}.~This shift transforms conventional computer vision from a static diagnostic tool into a dynamic algorithm that makes them uniquely suited for real-time robotic operations such as targeted spraying, where adaptability to novel scenarios, such as unexpected lighting conditions, unfamiliar crop varieties, or occluded targets, is essential.

Beyond perception tasks, the integration of vision-based FMs with reinforcement learning (RL), adaptive learning (AL), imitation learning (IL), and robotics represents the next frontier in agricultural automation~\cite{khosravi2025optimizing,kim2025autonomous,li2025enhanced}.~RL allows autonomous systems to optimize actions through continuous feedback, a critical feature in tasks such as precision spraying, robotic scouting, or unmanned aerial system (UAS)-based disease surveillance~\cite{van2025uav}. Adaptive learning extends this further by enabling models to evolve with changing environmental conditions, crop growth stages, or disease dynamics, ensuring sustained accuracy in dynamic field settings~\cite{abdalla2024adaptive,xu2025knowledge}.~When embedded within robotic platforms, such as ground vehicles or aerial drones, foundation models can empower autonomous systems to not only perceive but also reason and act in complex environments, bridging the gap between sensing and intelligent intervention. This coupling of large-scale pre-trained vision models with autonomous decision-making frameworks signals a paradigm shift toward fully integrated, self-improving agricultural robotics.

Despite their promise, significant challenges remain before FMs can be fully leveraged in agricultural domain, specifically for crop disease and pest management.~\thl{Apricot}{First}, there is a lack of large-scale, domain-specific benchmarking to assess their generalization across diverse crop species and field conditions~\cite{liu2024multimodal}.~\thl{Apricot}{Second}, identifying crop diseases is inherently complex and often requires the intervention of expert plant pathologists. Since it involves a thorough understanding of literature often combined with laboratory validation, the application of FMs can be challenging. For instance, when two disease symptoms appear very similar, FMs may struggle to reason correctly about their causes and differentiate between them.~In such cases, reasoning models may also suffer from an ``overthinking'' phenomenon, generating redundant outputs even after identifying the correct result~\cite{shojaee2025illusion}.~\thl{Apricot}{Third}, the integration of reasoning frameworks, RL, and adaptive mechanisms with FMs is still in its infancy, limiting their potential for real-time decision-making in autonomous systems~\cite{chen2025integrating,zaremehrjerdi2025towards}.~\thl{Apricot}{Finally}, issues surrounding data governance, model interpretability, and ethical deployment in agricultural settings require urgent attention~\cite{papagiannidis2025responsible}.~Against this backdrop, this review provides a timely synthesis of the current state of FMs in crop disease and pest management, examines their emerging applications in adaptive learning and robotics, and outlines promising directions for future research that may shape the trajectory of agricultural AI beyond 2025. Building on current developments in leveraging FMs for real-time, feedback-driven spraying systems, the article offer readers a comprehensive overview of advancements, identifies key challenges, and highlights opportunities for further exploration.~The specific contributions of this review are as follows:

\begin{enumerate}
    \item Examine current research trends to determine whether efforts are primarily focused on using large language models (LLMs) to synthesize extensive corpora of extension texts.
    \item Investigate the growing emphasis on integrating both vision and language modalities to enhance crop disease decision-making.
    \item Demonstrate how reasoning-based models are transforming paradigms, not only in vision-based systems but also through textual explanations, by engaging users in understanding why a model made a specific decision.
    \item Explore future directions in adaptive learning by first performing simulations within digital twin environments to test and refine models using reinforcement learning frameworks, and then developing and deploying practical, end-to-end systems that continuously adapt to real-world feedback.
\end{enumerate}

\section{Materials and Methods} \label{sec:problem}
\subsection{Comprehensive literature search}

\begin{figure*}[h]
\centering
    \hspace{0.5cm} \includegraphics[scale=0.55]{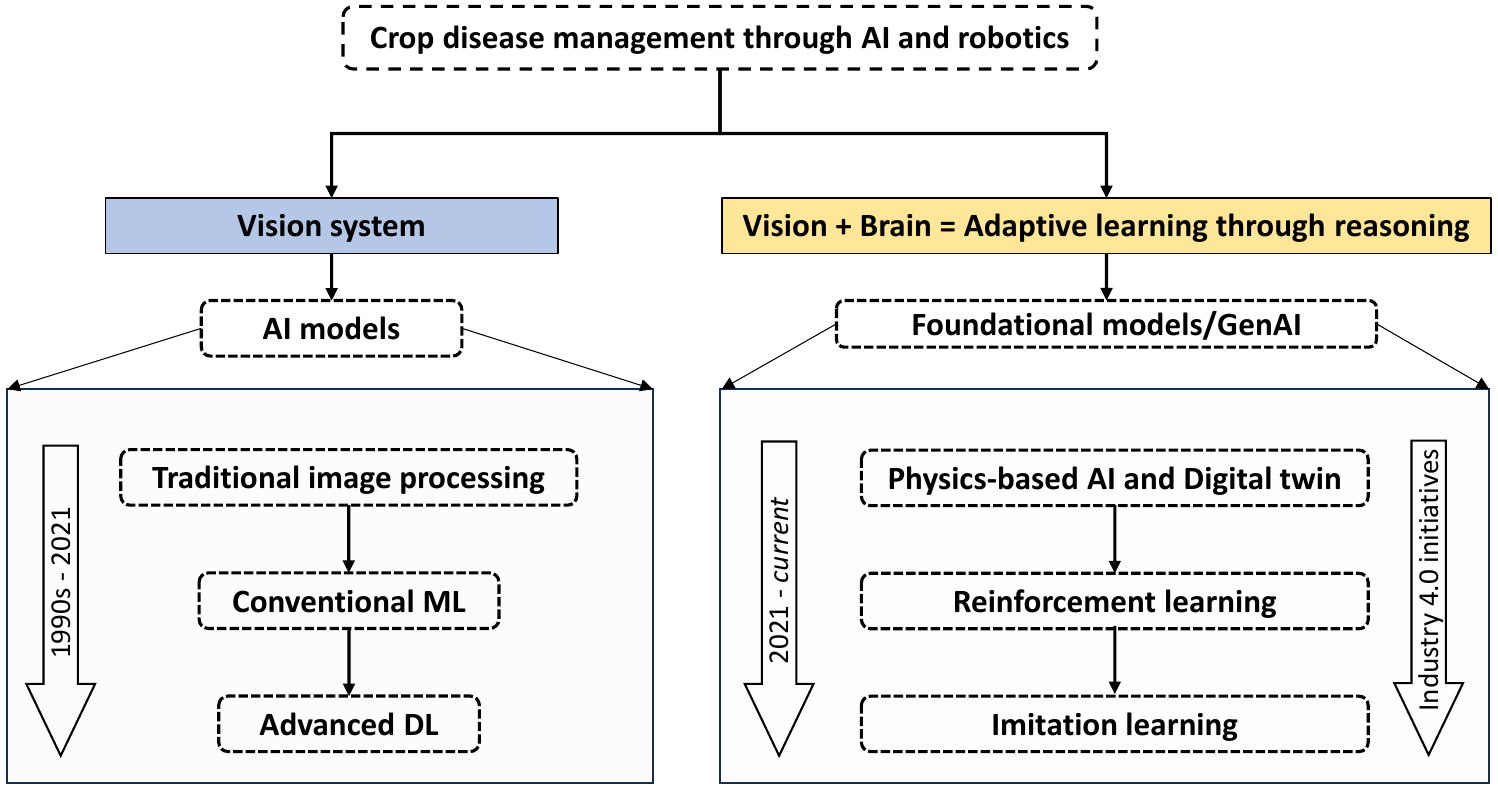}
     \caption{An overview of the review performed in this study with a perspective to focus on the individual steps taken to address site-specific disease management.}
     \label{fig1}
 \end{figure*}

The overall literature search and review analysis were performed for the last five years (2019-2024), focusing on the role of foundation models (FMs) in advancing site-specific disease management (SSDM). Beyond this central theme, the broader aim was to examine how these models could evolve into feedback-driven reasoning frameworks by integrating digital twins (DT), reinforcement learning (RL), and robotics to further support SSDM. Therefore, the analysis was split into two categories:~(a) vision system, and (b) vision + brain for adaptive learning and reasoning in real-time applications (Fig.~\ref{fig1}). 

For a comprehensive literature search, two academic databases were selected,~\href{https://www.sciencedirect.com/}{ScienceDirect} and \href{https://www.elsevier.com/products/scopus?dgcid=RN_AGCM_Sourced_300005030}{Scopus}.~However, to critically provide assessment on the current technologies as per the industry 4.0 initiatives, autonomous robots for site-specific disease management was also included~\cite{abbasi2022digitization}.~As part of advanced literature search, several keywords were used in conjunction with multiple Boolean operators, ``AND,'' and ``OR.''~Additionally, retrieved papers were adjudicated based on multiple screening criteria:~(a) crop disease identification with an aim to address site-specific disease management,~(b) peer-reviewed,~(c) English language,~(d) only research articles,~and~(e) duplicate papers across databases.~Table~\ref{tab1} reports the difference between papers retrieved and reported after literature search review process.

\begin{figure}[h]
    %\centering
    \includegraphics[scale=0.5]{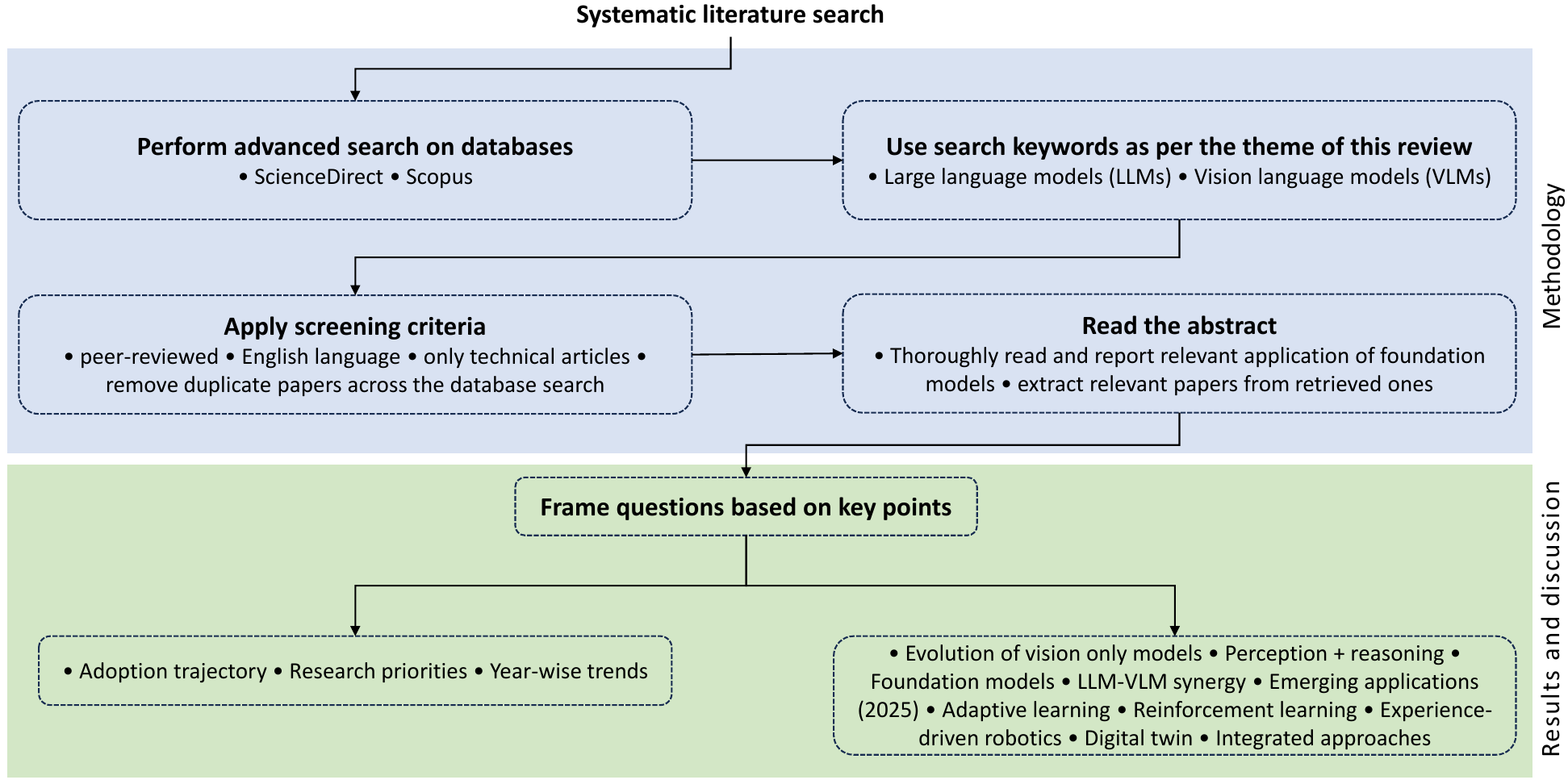}
    \caption{This review presents a systematic literature analysis focused on foundation models and their extended applications in site-specific crop disease management. The box on the left illustrates the employed methodology, while the box on the right highlights key questions derived from the retrieved research articles.}
    \label{screening}
\end{figure}

\begin{table*}[h]
\centering
\small
   \caption{Selected relevant publications on leveraging large language models and vision language models for crop disease and pest management.}
   \renewcommand{\arraystretch}{1}
   %\Large
      \resizebox{\textwidth}{!}{%
      \begin{tabular}{p{6cm}|>{\centering\arraybackslash}p{4cm}|>{\centering\arraybackslash}p{4cm}}
      \toprule
      Databases & Retrieved articles & Relevant articles \\ 
      \midrule
      \multicolumn{3}{p{15cm}}{\centering\arraybackslash\texttt{\scriptsize{(``large language models'' OR ``LLM'' OR ``foundation models'') AND (``plant disease'' OR ``crop disease'' OR ``pest management'' OR ``agricultural disease'')}}} \\
            \midrule
 ScienceDirect & 49 & 6 \\
 %IEEE \textit{Xplore} & 6 &  \\
 Scopus & 141 & 5 \\
 \midrule
      \multicolumn{3}{p{15cm}}{\centering\arraybackslash\texttt{\scriptsize{(``vision language model'' OR ``VLM'' OR ``foundation models'' OR ``multi-modal'') AND (``plant disease'' OR ``crop disease'' OR ``pest management'' OR ``agricultural disease'')}}} \\
      \midrule
      ScienceDirect & 107 & 16 \\
%      %IEEE \textit{Xplore} && \\
      Scopus & 569 & 11 \\
      \bottomrule
      \end{tabular}
      }
      \begin{tablenotes}
           \item \hspace{-0.65cm} \footnotesize{\textit{Note: Studies included were published between 2019 and 2024.}}
      \end{tablenotes}
      \label{tab1}
% \end{threeparttable}
 \end{table*}

\subsection{Article screening criteria and framing questions derived from key points}

The overall search and screening criteria was further subdivided into two categories based on the overall theme of research performed in crop disease management in precision agriculture (Fig.~\ref{screening}).~These were:~(a) large-language models (LLMs) as smart advisors for crop diseases, and (b) large vision models (VLMs) for smart crop detection and textual understanding/reasoning.~In the first category, technical research articles that focused on using LLMs to synthesize texts, such as extension articles or prescriptions, and to develop Q\&A platforms were selected.~In the second category, articles that leveraged VLMs, such as DINO, CLIP, or GPT, and combined multimodal data such as images, text, or sensor readings (structured data), were included.~This categorization was carried out to report results and discussion on each approach to crop disease management in precision agriculture. To accomplish this, two key phrases in the advanced search section of the databases were used.~For LLMS:~\texttt{(``large language models'' OR ``LLM'' OR ``foundation models'') AND (``plant disease'' OR ``crop disease'' OR ``pest management'' OR ``agricultural disease'')}, and for VLMs: \texttt{(``vision language model'' OR ``VLM'' OR ``foundation models'' OR ``multi-modal'') AND (``plant disease'' OR ``crop disease'' OR ``pest management'' OR ``agricultural disease'')}.~Based on the retrieved articles, a few key points were identified and used to frame two important questions for results analysis and eight additional questions for discussion (see Fig.~\ref{screening}).~These questions were: 

\begin{enumerate}
    \item[a.] What does the overall adoption trajectory of foundation models in crop disease research reveal about the pace and direction of this emerging field? (Sec.~\ref{sec3.2})
    \item[b.] What does the year-wise distribution of large language models (LLMs) and vision language models (VLMs)-based studies suggest about evolving research priorities in agricultural disease management? (Sec.~\ref{sec3.3})
    \item[c.] How does the increasing complexity of image acquisition sensors influence the evolution from traditional image processing to advanced deep learning? (Sec.~\ref{visionsystem})
    \item[d.] How are foundation models transforming vision systems into `vision + brain' frameworks that both perceive and reason about crop diseases? (Sec.~\ref{sec4.2})
    \item[e.] What are the emerging trends in the adoption and application of foundation models for crop disease management in the first half of 2025? (Sec.~\ref{4.3})
    \item[f.] How are reinforcement learning, adaptive learning, and experience-driven approaches being applied in agricultural robotics for crop disease management? (Sec.~\ref{sec4.4.1}) 
    \item[g.] How are digital twin technologies being leveraged for real-time monitoring and decision-making in crop disease management? (Sec.~\ref{Sec4.4.2})
    \item[h.] What are the benefits of combining reinforcement learning with digital twins in disease management? (Sec.~\ref{4.4.3})
\end{enumerate}

\section{Search Results}

\subsection{What does the overall adoption trajectory of foundation models in crop disease research reveal about the pace and direction of this emerging field?}\label{sec3.2}

\begin{figure}[h]
    \centering
    \includegraphics[scale=0.65]{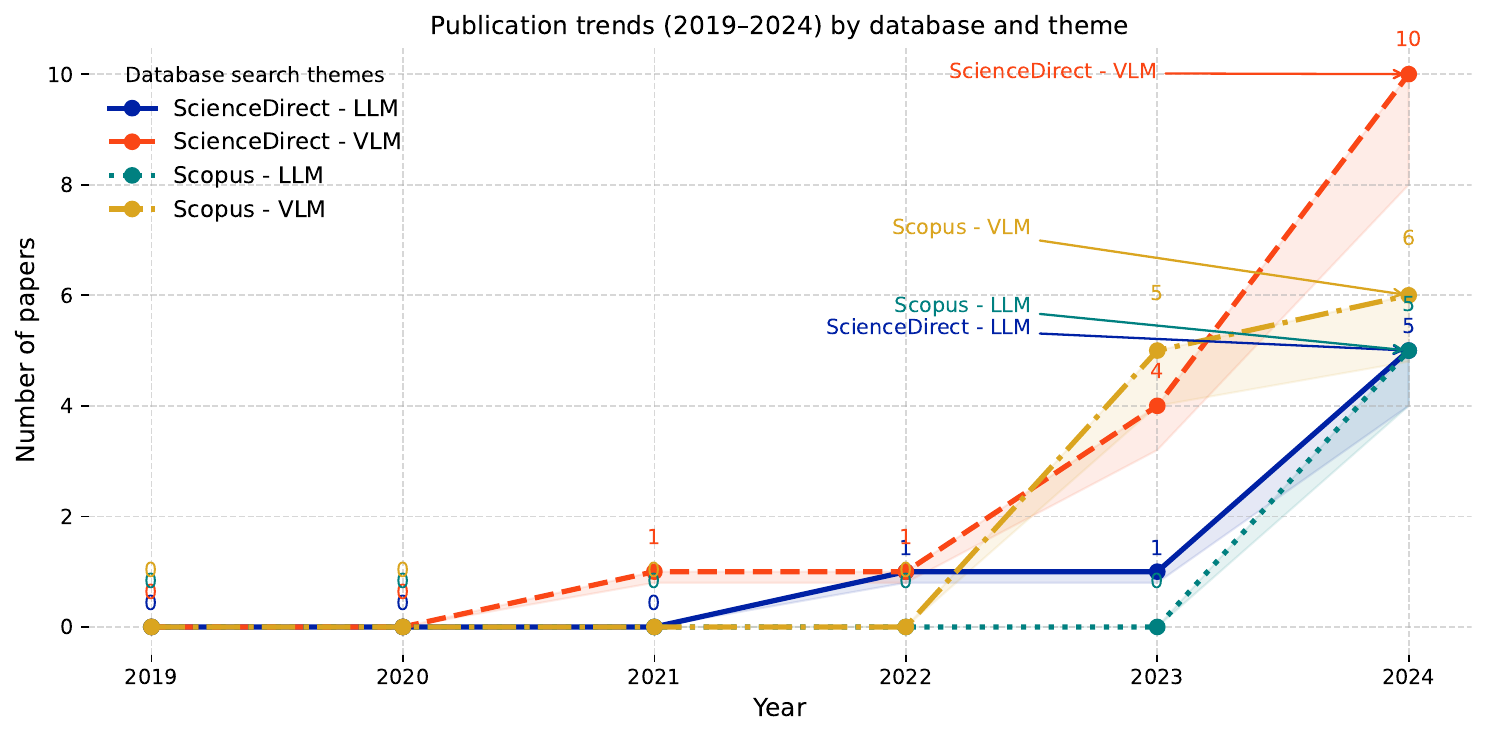}
    \caption{Publication trends from 2019 to 2024 for large language models (LLM) and visual language models (VLM) across ScienceDirect and Scopus databases.}
    \label{pubtrend}
\end{figure}

Table~\ref{tab1} summarizes the results of a literature search across two major databases to identify recent studies leveraging LLMs and VLMss for crop disease management and pest control. The first query, which combined LLM-related keywords with terms for plant and crop disease or pest management, yielded 49 articles from ScienceDirect and 141 from Scopus, but only a small subset directly relevant to the topic (Table~\ref{tab1}).~Although the literature search covered a five-year period, the majority of relevant articles were published in 2023 and 2024.~This indicates that, while LLMs such as OpenAI's GPT, released in 2018, are considered FMs, their application for synthesizing agricultural texts and crop disease knowledge only began to gain attraction around late 2022~\cite{radford2018improving} (Fig.~\ref{pubtrend}).~These findings highlight a significant gap between the rapidly expanding body of AI research and its direct application to agricultural-centered crop disease and pest management, suggesting that although foundational AI methods exist, their integration into practical agronomic systems remains limited.

\subsection{What does the year-wise distribution of large language models (LLMs) and vision language models (VLMs)-based studies suggest about evolving research priorities in agricultural disease management?} \label{sec3.3}

The publication trends from 2019 to 2024 reveal an increase in research activity on LLMs and VLMs across ScienceDirect and Scopus databases (Fig.~\ref{pubtrend}).~For LLM-related publications, ScienceDirect shows growth from 0-1 papers per year during 2019-2022 to 5 papers in 2024, indicating a fivefold increase, while Scopus records a rise from 0 papers in 2019-2022 to 6 papers in 2024, reflecting a similar trajectory. Based on the search term \texttt{(``vision language model'' OR ``foundational model'') AND (``plant disease'' OR ``crop disease'')}, over 107 and 569 research articles were retrieved in ScienceDirect and Scopus databases, respectively. Out of these VLM-centered research demonstrates an even sharper rise as ScienceDirect publications increase from 0-1 papers in 2019-2021 to 10 papers in 2024, representing a tenfold increase, highlighting the rapidly growing interest in visual-language integration in crop disease diagnostic studies. Scopus VLM papers also increase from 0-5 papers between 2019-2023 to 6 papers in 2024, showing steady growth. Notably, the largest annual jump occurs in ScienceDirect VLM papers between 2023 and 2024, suggesting a recent surge in research activity.~Overall, the trends indicate that VLM research is attracting more attention than LLM research, particularly through the use of multi-modal datasets and vision-based FMs. The relative growth in both databases reflects the broader adoption of FMs for applications, such as prompt-based reasoning to interpret disease symptoms and large-scale text synthesis for developing Q\&A platforms.~These patterns indicate a growing focus on integrating textual and visual data, with VLMs emerging as a central research area. 

\section{Discussion} \label{sec:algorithms}

\subsection{How does the increasing complexity of image acquisition sensors influence the evolution from traditional image processing to advanced deep learning?} \label{visionsystem}

\begin{figure*}[h]
    \centering
    \includegraphics[scale=0.5]{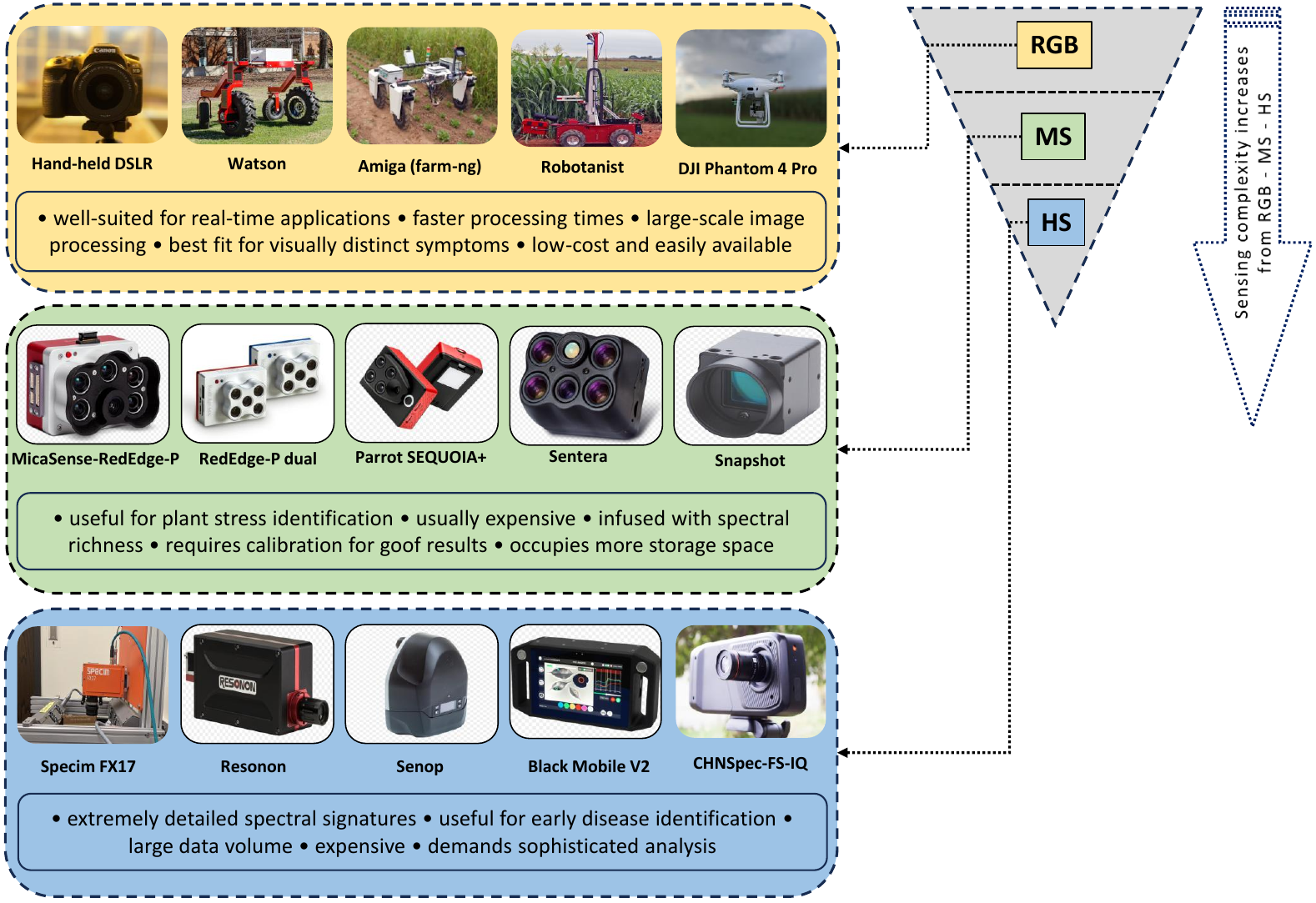}
    \caption{Different categories of sensing systems and platforms used in disease image acquisition.~The inverted triangle includes: RGB, multispectral (MS), and hyperspectral (HS) sensors where its application in sensing disease information increases from RGB to HS.}
    \label{fig2}
\end{figure*}

Current sensing platforms are the foundation for large-scale disease identification technologies, ranging from autonomous ground robots to tractor-mounted smart sprayers.~Over the years these sensors have been made more compact in space requirements without compromising on its high-resolution image acquisition capabilities.~Three categories of sensing system utilized largely for disease identification are: RGB, multi-spectral (MS), and hyperspectral (HS) (Fig.~\ref{fig2}).~For image acquisition in in-field settings, these sensors are either mounted on custom built chassis or commercialized autonomous robotic platforms.~Among all the three categories, RGB sensors have played a vital role in accomplishing disease identification to target spraying tasks~\cite{kim2021short,fu2022maize,davidson2022ndvi}.~While RGB sensors work on a very small wavelength (400-700 nm), thermal sensors are also used to scout for diseases and is mostly used to map wavelengths between 8,000-14,000 nm~\cite{zhu2024intelligent, upadhyay2025deep}.~These sensors does not have a very wide range of applications especially considering its useful assessment in extracting more information from crops.~Figure~\ref{fig2} showcases the inverted triangle which demonstrates the increase in complexity and sensor application from RGB to HS.~For instance, RGB sensors are a best fit to classify diseases that are visually distinct.~This use case also favors its real-time applications ability as the sensor does not involve detailed scanning of crops in order extract information in multiple bands, unlike MS and HS sensors.~As sensor complexity increases from RGB to HS sensors, so does the volume and dimensionality of data captured, more advanced data processing and analytical solutions.~This shift has enabled AI models to process large, complex datasets for timely and accurate disease diagnosis.

The integration of AI models, particularly those trained on numerous examples of large-scale disease datasets, has significantly advanced the ability to monitor crops in real-time for disease threats.~Between 2010 and 2017, most studies relied on leveraging either traditional image processing or conventional machine learning (ML) model based on handcrafted features to perform disease classification tasks in crops such as tomato, grapes, and maize~\cite{camargo2009image,packa2015morpho,sannakki2013diagnosis,jadhav2016grading} (Fig.~\ref{fig1}).~Although these approaches were limited in scale and real-time capability, it provided foundational research breakthroughs in crop disease identification. Building on this foundation, deep learning (DL) techniques, particularly, convolutional neural networks (CNNs), began to dominate the field post 2017.~These models significantly outperformed traditional methods by automatically extracting high-level features from raw image data, eliminating the need for manual feature engineering. Because of this attribute, these models come in different sizes where optimizing and integrating these on sUAS, edge computing devices, and robotic platforms have become possible. 

On the AI model architecture side, the first wave of CNN-based models for crop disease class emerged with transfer learning (2012-2015), where pretrained CNNs such as AlexNet, VGG, and ResNet were adapted to crop disease datasets, significantly improving classification accuracy despite limited data~\cite{bedi202118,brahimi2017deep}.~The second wave (2016-2019) saw real-time object detection models like Faster R-CNN, YOLO, and SSD applied to both identify and localize multiple diseases within crop canopies, enabling targeted spraying strategies~\cite{chen2020using, fuentes2017robust}. More recently, starting around 2020 onwards, transformer-based architectures and hybrid CNN-transformer models have been introduced, leveraging attention mechanisms to capture global context and fine-grained spatial relationships, further improving disease recognition and robustness across varying field conditions~\cite{thakur2022explainable,tonmoy2025mobileplantvit}.~Together, these advances in sensor modalities and learning paradigms have laid the foundation for next-generation precision agriculture systems capable of real-time disease monitoring and autonomous spraying.~Building on these advances, the rise of FMs marks the next leap, unifying multi-modal sensor data and learning methods into scalable, generalizable frameworks for agricultural disease management, capabilities not possible a few years ago.

\subsection{How are foundation models transforming vision systems into `vision + brain' frameworks that both perceive and reason about crop diseases?} \label{sec4.2}

\begin{figure}[h]
    \centering
    \includegraphics[scale=0.51]{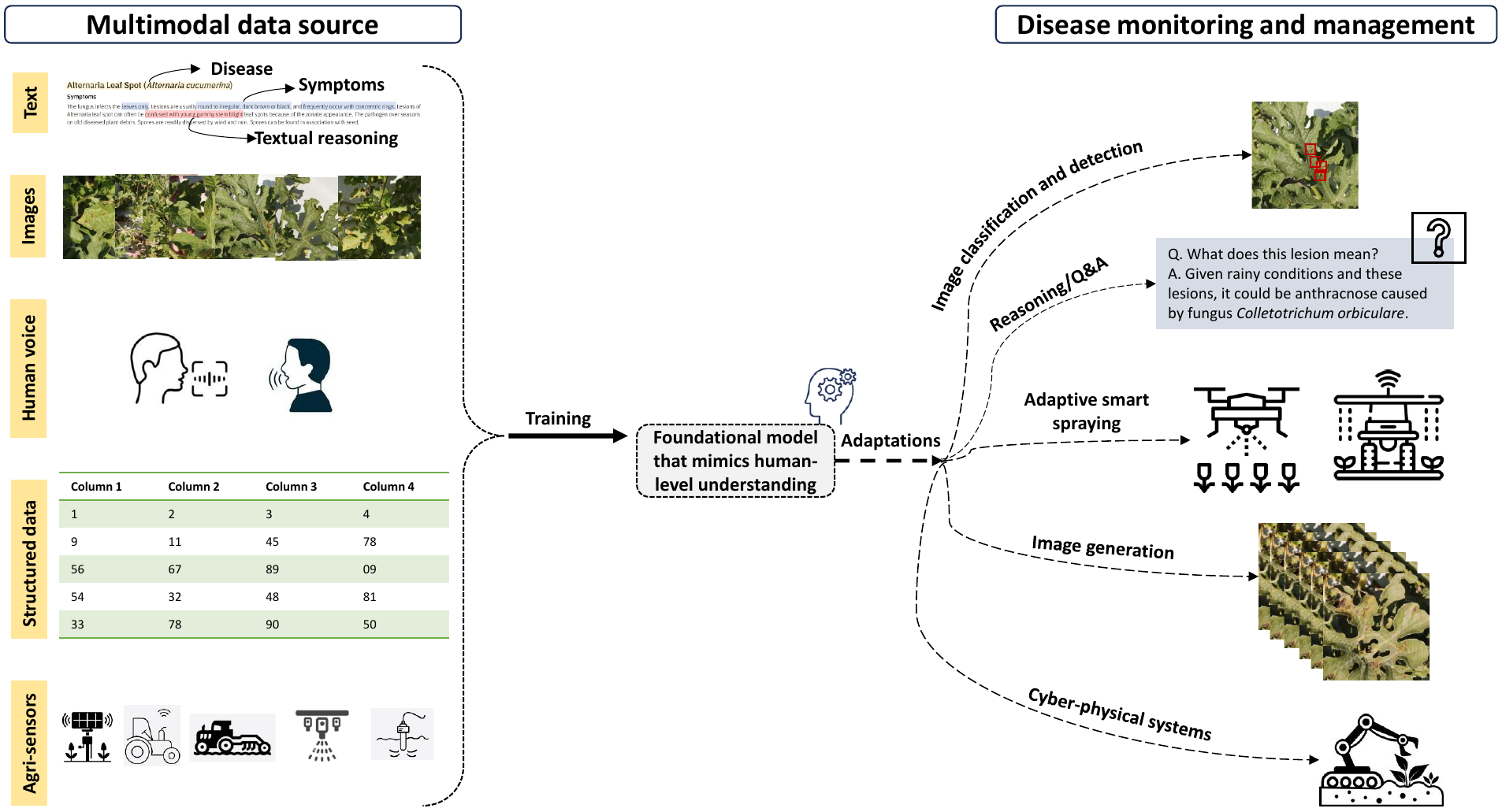}
    \caption{Overview of a foundational model-based framework for crop disease diagnostics and management.~Multimodal data sources including images, textual descriptions, structured dataset, and agri-sensors data. On the right, diverse downstream applications such as disease classification, image generation, smart spraying, question answering, and robotic navigation are enabled by the model's reasoning and generalization capabilities.}
    \label{foundationalmod}
\end{figure}

As discussed in Section~\ref{visionsystem}, traditional ML or advanced DL algorithms were mostly trained on one format of dataset, either images or structured dataset, and not both. On the contrary, FMs have the ability to be trained on a very diverse set of multi-modal data in the context of managing crop diseases~\cite{nvidiaFoundationModels}.~For instance, FMs trained on millions of images, weather pattern, soil data, and agricultural extension records (or texts) can understand crop diseases visually, interpret historical disease reports, and even process sensor data for possible disease outbreaks.~This could be a game-changing application for growers and farmers, who not only wants to identify a particular disease with high accuracy, but also receive an explanation, suggested treatments, or an explanation how a particular disease could spread under diverse weather conditions.~What makes a FM powerful is their ability to adapt to dynamic in-field conditions.~A model originally trained on wheat crop diseases in India can, with minimal adaptation, can also help a strawberry grower in Florida.~Therefore, a ``true'' FM is not defined by its architecture, but by its training scale, generalizability, and transferability across tasks (Fig.~\ref{foundationalmod}).~A model can only be considered ``foundational'' if it satisfies the following criteria: (a) it is pre-trained on a massive and diverse dataset that spans multiple domains, (b) adaptable to suit multiple downstream tasks pertaining to detection, segmentation or even reasoning, achieved through fine-tuning or few-shot learning, (c) supports cross-modal or multimodal inputs, as seen in GPT-4~\cite{achiam2023gpt}, CLIP~\cite{radford2021learning}, and Flamingo~\cite{alayrac2022flamingo}.~Without fulfilling these criteria, even the most powerful model will be accurately described as a task-specific deep learning model, not a foundational one.   

In the context of crop disease management in precision agriculture applications, the application of FMs could be integrated with multiple aspects of data modalities: (a) language, (b) vision, and (c) vision-language-action (VLA).~These modalities represent different layers of how FMs can perceive, interpret, and act upon complex agricultural environments. Language-based models enable interpretation of extension articles, scientific literature, or even grower-centered queries in natural language.~Whereas, vision-based models focus of extracting relevant disease-specific features, such as lesions and canopy structure, from drones or ground-robots.~The VLA integration extends these capabilities by extending reasoning and perception-based decision-making, allowing the system to act autonomously in various tasks pertaining to spraying or alerting based on contextual understanding.~Together, these modalities enable a holistic, intelligent framework for site-specific disease management, which was not possible a few years ago.

\subsubsection{Large language models (LLMs) as smart advisors for crop diseases}

LLMs are emerging as powerful ``smart advisors'' for crop disease management by synthesizing scientific literature, extension materials, and local guidelines into concise, actionable texts that can be understood by growers, extension educators, and farmers, alike. For instance, generative chatbots and retrieval augmented system (RAG)-style systems have been deployed to deliver localized recommendations, answer farmer questions in natural language, and turn complex research findings into step-by-step disease management advice~\href{https://digitalgreen.org/farmer-chat/}{(Link)}.~Researchers have shown that LLMs can summarize diseases and pests life cycles, generate treatment protocols, produce multilingual extension information, and generate scripted troubleshooting flows for field technicians, a task that directly reduce the time between observation and intervention~\cite{Zhao2024,tzachor2023large}.

Multi-modal extensions of LLMs that combine combine image encoders with text decoders are beginning to link visual disease diagnosis with text generation, allowing a user to upload a leaf photo and receive both a likely diagnosis and an extension-style recommendation.~In practice, these capabilities can help small-scale farmers and extension educators by: (1) generating localized management plans (in local languages) from regional weather/phenology data, (2) converting research papers into farmer-friendly protocols, and (3) producing training modules and quiz material for extension workshops.~A few recent studies have leveraged LLMs to synthesize large corpora of agricultural texts and generate targeted solutions for specific problems.~For instance, combining LLMs with agricultural knowledge graphs enables efficient reasoning over complex disease symptoms, providing accurate plant disease detection guidance~\cite{Zhao2024}.~LLMs can also automate the synthesis of pest control literature, reducing the burden of manual review and rapidly offering actionable insights~\cite{Scheepens20241261}. Their broader potential in crop production includes summarizing vast amounts of literature and supporting decision-making~\cite{Kuska2024}.~Beyond text synthesis, integrating LLMs with sensor data allows plant health monitoring to be queried and explained in natural language~\cite{Ahir2024178}.~At the systemic level, LLMs can enhance agricultural extension services, providing farmers with timely and contextually accurate recommendations on disease and pest management~\cite{tzachor2023large}. Table~\ref{tab2} summarizes state-of-the-art LLM architectures used to synthesize large-scale information with respect to crop diseases and pest management in precision agriculture research.
%\lipsum[4] See Section \ref{sec:headings}.

\subsubsection{Vision language models (VLMs) for crop disease detection through reasoning} \label{VLM}

Vision language models (VLMs) for crop disease diagnosis offer transformative approach to plant health monitoring and disease classification. These models are built on architectures like Vision Transformers (ViT) or self-supervised frameworks, such as DINO, MAE, or CLIP, are pre-trained on massive, diverse image datasets and can generalize across a wide range of visual task. While ViTs or other similar architectures are commonly used in plant disease diagnosis tasks, it is important to note that their use alone does not qualify a model as foundational.~For instance, the application of ViT to accurately detect and classify Java Plum leaf disease on limited samples of six classes does not qualify to be a FM for crop disease diagnosis~\cite{bhowmik2024customised}. Another study by~\cite{thakur2023vision} used over 40,000 images from various open-access platforms

\begin{landscape}   % start landscape page
{\small % small font for readability
\begin{longtable}{>{\raggedright\arraybackslash}p{5.2cm}|
                  >{\raggedright\arraybackslash}p{3.5cm}|
                  >{\raggedright\arraybackslash}p{5cm}|
                  >{\raggedright\arraybackslash}p{5cm}|
                  >{\raggedright\arraybackslash}p{3cm}}
\caption{Summary of recent studies applying large language models (LLMs) for agricultural text synthesis, disease management, and extension services.} 
\label{tab2} \\
\toprule
Approach/Models & Open-access source & Use cases & Journal & References \\ 
\midrule
\endfirsthead

\multicolumn{5}{c}{{\bfseries Table \thetable\ (continued)}} \\
\toprule
Approach/Models & Open-access source & Use cases & Journal & References \\
\midrule
\endhead

\bottomrule
\endfoot

% ====================== 2024 ======================
\multicolumn{5}{c}{\textbf{2024}} \\
\midrule
Used LLM with Agricultural Knowledge Graphs (KGs), Graph Neural Networks (GNNs) & Name of the model not specified & Plant disease diagnosis systems, reasoning over symptom descriptions, linking textual disease corpora with structured knowledge & MDPI \textit{(agriculture)} & \cite{Zhao2024} \\ 

Used GPT-4 (OpenAI API) for automated literature synthesis on pest controllers & Proprietary (OpenAI), not open-source & Automating systematic reviews in pest control, reducing expert workload in literature mining & \textit{Methods in Ecology and Evolution} & \cite{Scheepens20241261} \\

Experimented with GPT-based language models to process sensor + text queries & Proprietary (OpenAI), not open-source & Query-based plant health monitoring. Example: ``Why is my leaf showing yellowing?'' explained using sensor readings + LLM reasoning & \textit{International Journal of Computer Applications in Technology} & \cite{Ahir2024178} \\

Question-answering systems in agriculture (covering corpora, knowledge graphs, large language models like GPT-4) & --- & Processing agricultural queries, including plant disease diagnosis, pest identification, and disease control, using Q\&A systems for intelligent production and sustainable management & \textit{Resources, Conservation and Recycling} & \cite{yang2024application} \\

GlyReShot (glyph-aware few-shot Chinese agricultural named entity recognition integrating a lightweight GROM module and training-free label refinement strategy) & --- & Recognizing entities like diseases, crop, pest, and drug in Chinese agricultural text, including improved recognition of plant disease entities under scarce labeled data conditions & \textit{Heliyon} & \cite{liu2024glyreshot} \\

RAG chatbot combining hybrid DeiT + VGG16 CNN model for medicinal plant identification and insights, incorporates Retrieval-Augmented Generation and explainable AI & Not explicitly available & Identifying medicinal plants (via images) and generating bilingual insights (Nepali \& English), including health benefits, cultivation tips, using hybrid deep learning + RAG & \textit{Telematics and Informatics Report} & \cite{paneru2024leveraging} \\

Agricultural Knowledge Graph (AGKG): integrates NLP and deep learning with LLMs to automatically extract agricultural entities and construct a knowledge graph for engineering technology applications & Not specified & Agricultural entity retrieval and Q\&A via domain-specific AGKG built from Internet data & \textit{Displays} & \cite{wang2024digital} \\

% ====================== 2023 ======================
\midrule
\multicolumn{5}{c}{\textbf{2023}} \\
\midrule
Used GPT-3.5 for agricultural extension services & Proprietary (OpenAI), but can be replicated with open Hugging Face models & Farmer advisory chatbots, pest and disease diagnosis, local language extension support & \textit{Nature Food} & \cite{tzachor2023large} \\

ChatAgri (ChatGPT-based agricultural text classification using prompt engineering strategies across languages) & \href{https://github.com/albert-jin/agricultural_textual_classification_ChatGPT}{GitHub link} & Cross-linguistic agricultural news text classification; few-shot and prompt-based classification using GPT-3.5 and GPT-4 & \textit{Neurocomputing} & \cite{zhao2023chatagri} \\

% ====================== 2022 ======================
%\midrule
\multicolumn{5}{c}{\textbf{2022}} \\ 
\midrule
AgriBERT (BERT-based, pretrained on 300M agri-food tokens, knowledge-infused with FoodOn/Wikidata) & --- & Semantic matching of food descriptions to nutrition entries, cuisine classification, and agricultural NLP tasks & International Joint Conferences on Artificial Intelligence (IJCAI) 2022 & \cite{rezayi2022agribert} \\

% ====================== 2018 ======================
\midrule
\multicolumn{5}{c}{\textbf{2018}} \\ 
\midrule
Original GPT pre-training paper (OpenAI) & Not open-source at the time, later GPT-2/3 derivatives inspired open-source releases & Laid foundation work for all the later agricultural LLM applications & OpenAI & \cite{radford2018improving} \\

\end{longtable}
} % end small font
\end{landscape}

\noindent to integrate vision transformers with CNNs to address disease classification in multiple crops. Although the number of dataset used in this research was moderately large-scale, it did not involve any cross-domain adaptations, a key requirement for a model to be called as a FM. Therefore, a FM should go beyond classification task and demonstrate broad generalization across crops, sensor modalities, and disease types, thereby highlighting its versatility and cross-domain adaptations.

In a recent study, the authors proposed a vision-language framework that uses texts prompts to guide a vision model for disease anomaly detection~\cite{dong2024visual}. Another notable work from 2023 integrated the You Only Look Once (YOLO) model with GPT-guided textual understanding to generate crop diagnostics report~\cite{qing2023gpt}.~Similarly,~\cite{zhang2024visual} employed the Segment Anything Model (SAM) to first isolate wheat diseases and then applied a reasoning chain framework to generate well-structured diagnostic explanations.~In another study, a few-shot learning technique was applied to train the PlantCaFo model for disease recognition~\cite{jiang2025plantcafo}. With this growing interest in this space, it is evident that FMs will play a dominant role in future agricultural AI research, particularly in supporting multi-modal outputs, not just classification, but explanation, generation, and recommendations with reasoning. This mirrors trends observed in the early evolution of DL, where CNN-based approaches initially dominated before being surpassed by more generalized models.~Additionally, most of the existing studies mostly rely on pre-existing models, such as GPT, CLIP, or SAM, which are then fine-tuned on agricultural-centered texts.~Although these models can be adapted through one-shot or few-shot learning, their veracity remains limited when it comes to handling domain-specific terminology in plant pathology. Some existing domain-specific models (BERT-based models), such as AgriBERT, do exist, but these are relatively small-scale and are primarily tailored for natural language processing (NLP) tasks, rather than serving as full LLMs~\cite{rezayi2022agribert}. Table~\ref{tab3} summarizes the application of multi-modal VLMs specifically used to synthesize texts and present reasoning with respect to diseased leaf images.

\subsection{What are the emerging trends in the adoption and application of foundation models for crop disease management in the first half of 2025?} \label{4.3}

In 2025, crop disease and pest management using AI approaches is shifting from simple vision classifiers toward FM pipelines that are multi-modal and label-efficient.~VLMs, CLIP-style backbones, diffusion generators, and FM adapters are being integrated with various spectral and physiological information about crop diseases.~Additionally, FMs are being integrated with robotics to enable diagnosis, explanation, and action. The studies below highlight emerging trends in their application to SSDM in precision agriculture. These are: 

\begin{enumerate}
    \item[a.] \textbf{Domain-adapted VLMs replace task-specific CNNs:} The development of PlantCaFo
    demonstrated that FMs could be used for crop disease tasks through lightweight adapters and weight decomposition, enabling accurate classification even in few-shot scenarios~\cite{jiang2025plantcafo}.~In parallel, FMs, such as DINOv2 could be effectively repurposed for tasks such as crop disease recognition with minimal adaptation.~To enhance practicality and efficiency, the study employed strategies including linear probing, parameter-efficient fine-tuning using LoRA techniques, and knowledge distillation into smaller architectures like MobileNetV3~\cite{espejo2025foundation}.

    \item [b.] \textbf{Few-shot/zero-shot with prompt or adapter tuning becomes practical:} Further advancements in the field have also been fueled due to approaches such as one-shot and few -shot. For instance, Progressive Mixup Prompt Learning combined with CLIP Dynamic Calibration (CDC) introduced an innovative approach to unsupervised test-time domain adaptation, enabling models to generalize to novel disease datasets without relying on fully supervised retraining~\cite{chen2025enhancing}.~Complementing this, ChatLeafDisease demonstrates the effective integration of LLM chain-of-thought reasoning with curated disease descriptions, achieving training-free tomato disease classification that surpasses generic GPT and vision-language model baselines~\cite{pan2025chatleafdisease}.

    \item[c.] \textbf{Multi-modal fusion (image + language + spectral signatures) improves disease assessment and field robustness:}~Recent works in 2025 also highlights an merging trend where researchers are increasingly integrating VLMs with LLMs to not only detect crop diseases but also generate actionable prescriptions for farmers.~For instance, a DL-based computer vision model, YOLO, was integrated with an LLM to generate both crop detection and treatment recommendations~\cite{yan2025cdip}. Another work developed the SCOLD (Soft-target COntrastive Learning for Leaf Disease identification) model and trained it on over 186,000 image-caption pairs covering 97 unique concepts to address context-aware crop disease classification~\cite{quoc2025vision}.

    \item[d.] \textbf{Text-to-image-based synthetic generation of crop disease images:}~To gain more control over the generated images, researchers have started utilizing text prompts to describe specific disease symptoms such as lesion color, shape, or leaf texture. For instance, work by~\cite{rai2025phytosynth} compared three Stable Diffusion (SD) architectures, SDXL, SD3.5-medium, and SD3.5-large, to generate realistic looking images multiple diseases in watermelons (Fig.~\ref{VLM}). Their research found out that training as small as 36 real images could generate 500 (or more) synthetic images of varying environmental and dynamic backgrounds when trained on A100 Nvidia GPU for only 1.5 hours.

\end{enumerate}

\begin{landscape}   % start landscape page
{\small % small font for readability
\begin{longtable}{>{\raggedright\arraybackslash}p{3.5cm}|
                  >{\raggedright\arraybackslash}p{3.5cm}|
                  >{\raggedright\arraybackslash}p{3.5cm}|
                  >{\raggedright\arraybackslash}p{5.5cm}|
                  >{\raggedright\arraybackslash}p{3cm}|
                  >{\raggedright\arraybackslash}p{3.5cm}}
\caption{Summary of recent studies applying vision language models (VLMs) to combine both images and texts of crop diseases for context-based reasoning and learning.} 
\label{tab3} \\
\toprule
Models & Type & Open-access source & Use cases & Journal & References \\ 
\midrule
\endfirsthead

\multicolumn{6}{c}{{\bfseries Table \thetable\ (continued)}} \\
\toprule
Models & Type & Open-access source & Use cases & Journal & References \\ 
\midrule

\endhead

\bottomrule
\endfoot

\multicolumn{6}{c}{\textbf{2024}} \\
\midrule
DINOv2 & Vision model & \href{https://huggingface.co/docs/transformers/en/model_doc/dinov2}{Hugging Face} & Self-supervised feature extraction, clustering of disease symptoms & ScienceDirect & \cite{bai2024dinov2} \\

BLIP/BLIP-2 & Multi-modal model & \href{https://huggingface.co/docs/transformers/en/model_doc/blip-2}{Hugging Face} & Image captioning and visual reasoning for disease explanation & & \cite{liang2024dynamic} \\

LLaVA & Multi-modal model & \href{https://huggingface.co/docs/transformers/en/model_doc/llava}{Hugging Face} & Multi-modal reasoning for plant disease recognition & & \cite{xu2025agro} \\

SAM & VLM & \href{https://github.com/facebookresearch/segment-anything}{Facebook} & Wheat disease diagnosis through reasoning & Science Direct & \cite{zhang2024visual} \\

ViT + GPT-2 & VLM & \href{https://openai.com/index/gpt-2-1-5b-release/}{OpenAI}, \href{https://huggingface.co/docs/transformers/en/model_doc/vit}{Hugging Face} & Align plant disease phenotypes with trait descriptions & \textit{Plant Phenomics} & \cite{zhao2024plantext} \\

Inception-v4 + LSTM & VLM & --- & Aligning crop disease images with question embeddings & Plant Phenomics & \cite{zhao2024informed} \\
%DALL•E & Multi-modal model & \href{https://openai.com/index/dall-e/}{OpenAI} & Synthetic image generation for apple 

PDC-VLD & Multi-modal (vision + text) & No specific mention & Tomato leaf disease detection with unseen class generalization & \textit{Plant Phenomics} & \cite{li2024multi} \\

FHTW-Net & Vision language model (image-text retrieval) & \href{https://github.com/ZhouGuoXiong/FHTW-Net}{GitHub} (No specific mention of the model) & Retrieve matching text from a query image (and vice versa) for rice leaf disease descriptions & \textit{Plant Phenomics} & \cite{zhou2024precise} \\

A multi-modal Chinese model & Multi-modal prompting + texts & Not specified & Identifying diseases, pests, and control related entities in Chinese agricultural texts & \textit{Computers and Electronics in Agriculture} & \cite{zhang2024chinese} \\

ILCD (Informed-learning guided model of crop diseases) & Multi-modal visual question answering model & \href{https://github.com/SdustZYP/ILCD-master/tree/main}{GitHub} & The devleoped model addressed complex questions about crop disease stages and attributes & \textit{Plant Phenomics} & \cite{zhao2024informed} \\

PhenoTrait text description model ( GPT-4 and GPT-4o) & Multi-modal (image-to-text generation) & \href{https://plantext.samlab.cn/}{PlanText} & Novel model generates plant disease text from images & \textit{Plant Phenomics} & \cite{zhao2024plantext} \\

Multi-modal vegetable knowledge graph (No specific model name) & Multi-modal knowledge graph construction & Not specified & Used as a foundational tool to extract and process knowledge from text & \textit{Computers and Electronics in Agriculture} & \cite{lv2024veg} \\

Name not specified & Multi-modal foundation model & Not specified & Vision-language model with visual information to improve performance on fine-grained plant disease anomaly detection & \textit{Computers and Electronics in Agriculture}  & \cite{dong2024visual} \\

WDLM (Wheat disease language model) & Visual language model (VLM) & Not specified & Fine-tuning foundation models for wheat disease diagnosis & \textit{Computers and Electronics in Agriculture} & \cite{zhang2024visual} \\

PepperNet & Multi-modal vision-language model & Not specified & Detecting pepper diseases and pests in complex agricultural environments using natural language descriptions & \textit{Nature - Scientific Reports} & \cite{liu2024multimodal} \\

APD-229 (Agricultural pests and diseases) & Multi-modal (Textual-Visual) & Link is given but does not work & A multi-modal approach that uses text descriptions to guide an image recognition system for fine-grained classification & \textit{Multimedia Tools and Application} & \cite{wang2024apd} \\

Qwen-VL & Pre-trained VLM & \href{https://drive.google.com/drive/folders/1sl-nRDYGz9T4969QjS_nRnoZIfcREWOW}{Google Drive} (Only dataset, no specified model) & Used to generate meticulous text descriptions for disease images, which serve as prompt text for generating classifier weights & MDPI \textit{(sensors)} & \cite{zhou2024few} \\

Segment Anything Model (SAM) & Image segmentation & \href{https://segment-anything.com/}{SAM-Meta AI} & Indentifies and segments out all the suggested regions within the diseased leaf image & IEEE~\textit{Access} & \cite{moupojou2024segment} \\

Specific name not mentioned & Fine-tuning paradigms for out-of-distribution detection & \href{https://github.com/JiuqingDong/PDOOD}{GitHub} & multi-modal model (specifically, a visual-language model) was used to explore its effectiveness in OOD plant disease detection 
& Nature Scientific Reports & \cite{dong2024impact} \\

Visual answer model (VQA) & Multi-modal VQA & Not specified & A model designed to answer questions about fruit tree diseases by fusing image and Q\&A knowledge & \textit{Frontiers in Plant Science} & \cite{lan2023visual} \\

%%%%%%%%%%%%%%%%%%%%%%%%%%%%%%% 2023 %%%%%%%%%%%%%%%%%%%%%%%%%%%%%%%%%%%%%%

\midrule
\multicolumn{6}{c}{\textbf{2023}} \\
\midrule

ITLMLP (Image-to-text multi-modal model) & Vision language pre-training & Not specified & Few-shot learning to recognize cucumber diseases using a multi-modal approach that combines image, text, and label information & \textit{Computers and Electronics in Agriculture} & \cite{cao2023cucumber} \\

YOLO and GPT combined & Multi-modal model & \href{https://github.com/ultralytics/ultralytics}{GitHub}, \href{https://platform.openai.com/docs/models}{OpenAI} & Used for its deep logical reasoning capabilities to generate agricultural diagnostic reports & \textit{Computers and Electronics in Agriculture} & \cite{qing2023gpt} \\

ITF-WPI model & Cross-modal feature fusion model & \href{https://github.com/wemindful/Cross-modal-pest-Identifying}{GitHub} & The proposed model overcomes the challenges of complex agricultural backgrounds by using cross-modal data (images and text) for wolfberry pest identification & \textit{Computers and Electronics in Agriculture} & \cite{dai2023itf} \\

Specific name not mentioned & Neuro-symbolic AI (deep learning + knowledge graphs) & \href{https://github.com/Research-Tek/xai-cassava-agriculture/tree/master}{GitHub} & The main approach developed to improve prediction accuracy and generate user-level, understandable explanations for non-experts, such as farmers, by combining deep learning with domain knowledge & \textit{Expert Systems with Applications} & \cite{chhetri2023towards} \\

ShuffleNetV2 + TextCNN & Multi-modal model & Not specified & Models are used to extract textual features and semantic relationships from descriptive text & Nature - Scientific Report & \cite{qiu2023detection} \\

MMFGT (Multimodal fine-grained transformer) & Multi-modal transformer model & Not specified & A novel model for few-shot pest recognition that combines multi-modal information from images and text & MDPI \textit{(electronics)} & \cite{zhang2023multimodal} \\

ODP-Tranformer & Multi-modal (image-to-text generation + classification) & Not specified & A two-stage model proposed to interpret pest image classification results by generating captions, in addition to predicting labels & \textit{Computers and Electronics in Agriculture} & \cite{wang2023odp} \\

%%%%%%%%%%%%%%%%%%%%%%%%%%%%%%%%%%% 2022 %%%%%%%%%%%%%%%%%%%%%%%%%%%%%%%%%%%%%%

\midrule
\multicolumn{6}{c}{\textbf{2021}} \\
\midrule

ITK-Net (Image-text-Knowledge Network) & Multi-modal (Image-text-Knowledge) & Not specified & The primary model developed to identify common invasive diseases in tomato and cucumber by leveraging multimodal data and high-level domain knowledge & \textit{Computers and Electronics in Agriculture} & \cite{zhou2021crop} \\

\end{longtable}
} % end small font
\end{landscape}
    
%\newpage

    \begin{figure}
        %\centering
        \includegraphics[scale=0.5]{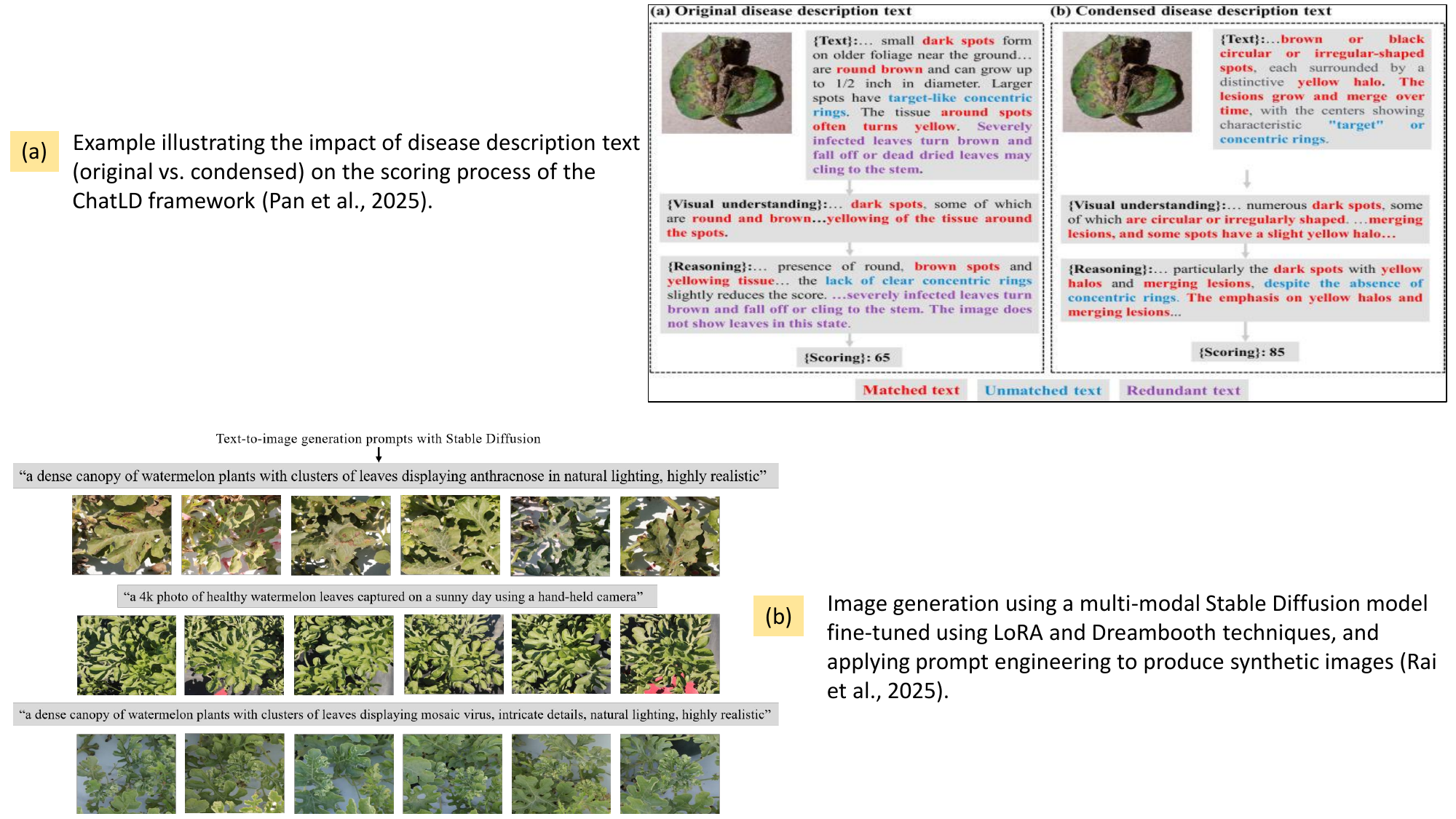}
        \caption{Example representing the role of foundation models (FMs) in the context of crop disease research, including vision-language integration, text-to-image generation, and multi-modal scoring frameworks.}
        \label{VLM}
    \end{figure}    

\subsection{Closing the loop in crop disease management by integrating reinforcement learning and digital twins in cyber-physical system}

A major future direction for precision crop disease and pest management is the transition from perception-driven systems to closed-loop, feedback-based learning frameworks (sense $\rightarrow$ simulate $\rightarrow$ decide $\rightarrow$ act $\rightarrow$ update).~Two complementary technologies are central to this vision: reinforcement learning (RL) and digital twins (DTs).~Reinforcement learning (RL) is an AI approach in which an agent discovers the best actions by interacting with its environment and learning from trial and error \cite{littman1996reinforcement}.~DTs, in contrast, are data-driven virtual replicas of real farm systems that stay synced with live sensor streams.~Integration of RL and DTs enable adaptive decision-making.~For example, RL can optimize robotic sprayers under the highly variable conditions found in the field, while DTs can simulate disease dynamics, crop growth, and management strategies under different scenarios.~When fed real- time sensor data and AI predictions, DTs can simulate the outcomes of proposed interventions, providing a safe environment for RL agents to refine policies before deployment.~This section surveys state-of-the-art applications of RL in agricultural robotics and spraying, the use of digital twins for crop monitoring and simulation, their combined role in disease-specific pesticide spraying, real- world case studies, and forward-looking perspectives on these emerging technologies.

\subsubsection{How are reinforcement learning, adaptive learning, and experience-driven approaches being applied in agricultural robotics for crop disease management?} \label{sec4.4.1}

\begin{figure}[h]
    \centering
    \includegraphics[scale=0.52]{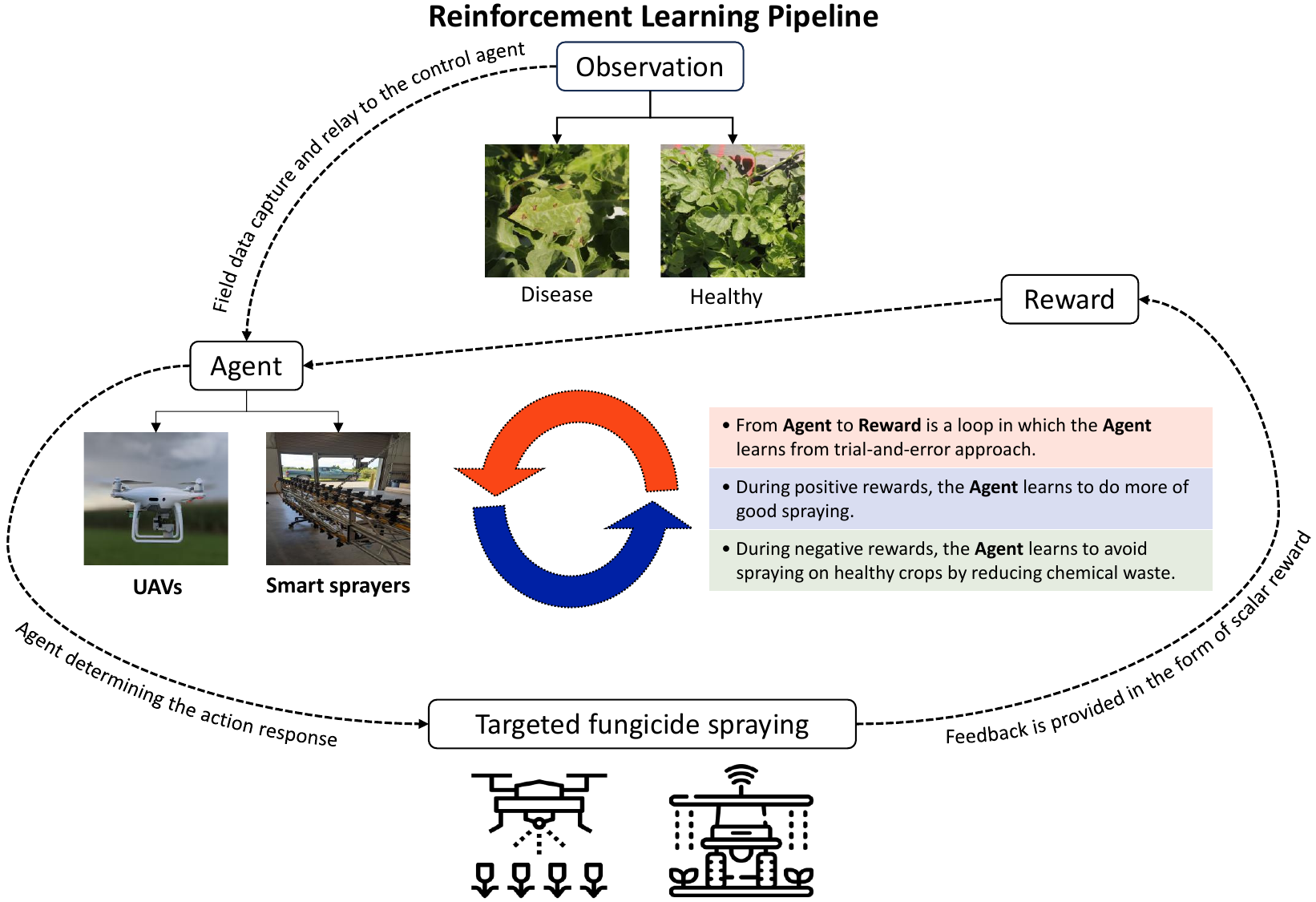}
    \caption{The overall agent and reward components of reinforcement learning for leveraging site-specific disease management in precision agriculture.}
    \label{reiforcement}
\end{figure}

Recent advancements in RL have demonstrated significant potential for transforming precision agriculture through autonomous technologies~\cite{wozniak2024recent,yepez2023mobile}.~These developments encompass a wide range of agricultural robotics applications, from autonomous navigation~\cite{goldenits2024current,yang2022intelligent} and hardware control~\cite{li2024peduncle,yandun2021reaching} to precision resource application systems~\cite{kelly2024assessing,siddique2019robust}.~Modern precision agriculture demands intelligent systems capable of making complex decisions in dynamic field environments, and RL provides a framework for developing autonomous technologies that can adapt to varying crop conditions, weather patterns, and field characteristics.~Precision spraying serves as one compelling example of this broader transformation toward intelligent, data-driven agricultural technologies, demonstrating how RL can optimize both operational efficiency and environmental sustainability~(Fig.~\ref{reiforcement}).~Although RL applications in precision agriculture are still emerging, existing examples illustrate a range of uses from low-level hardware control to high-level strategic decision-making, potentially addressing fundamental challenges of traditional rule-based crop management systems.

Unlike conventional rule-based agricultural systems, RL-enabled technologies learn from continuous field data and operational outcomes, enabling more sophisticated and context-aware decision-making processes.~For example,~\cite{hao2022adaptive} developed an RL-based decision system for UAV crop sprayers that dynamically adjusts spray volume, droplet size, and flight height based on real-time crop characteristics. Their actor-critic neural network, trained on comprehensive field experiment data, achieved a 14\% reduction in pesticide volume across wheat fields while maintaining equivalent pest control efficacy, demonstrating the potential for resource optimization across agricultural operations.~Other studies have used deep RL for path planning so unmanned aerial vehicle (UAV) only treat infected zones, avoiding healthy crops to save chemicals. In a recent simulation, a hierarchical RL agent was trained to navigate a field and spray only diseased plants, significantly improving yield protection while reducing unnecessary spraying~\cite{khosravi2025optimizing}.~Beyond a single UAV, RL has been extended to complex, multi-robot spraying scenarios.~In a simulation study,~\cite{farid2025multiple} explored how on-policy reinforcement-learning algorithms could coordinate multiple UAVs and ground vehicles for crop spraying.~One RL agent planned efficient coverage paths for several drones, while a second agent continually adjusted each UAV's position to counter simulated wind drift. In narrow field sections, the framework reassigned spraying duties to ground robots, demonstrating how hybrid air and ground agents might improve overall efficiency. 

RL for crop disease management is not limited to robotics hardware control. RL also offers a framework for decision support systems in pest control. By casting spraying schedules as a sequential decision problem, RL can learn when and how much to spray based on pest or disease population dynamics. For example,~\cite{siddique2019robust} addressed the problem of an orchard manager deciding whether to apply pesticides at each time step given pest population levels.~They formulated this scenario as an RL challenge and developed a reinforcement learning method that computes spraying policies likely to perform well despite noisy, imperfect pest data. The RL agent essentially learns an optimal integrated pest management strategy (when to spray or hold off) that maximizes long-term orchard health and yield. By incorporating uncertainty (via Bayesian modeling and robust optimization), their approach yields pest control policies that remain effective even with incomplete information.

\subsubsection{How are digital twin technologies being leveraged for real-time monitoring and decision-making in crop disease management?} \label{Sec4.4.2}

\begin{figure}[h]
    \centering
    \includegraphics[scale=0.47]{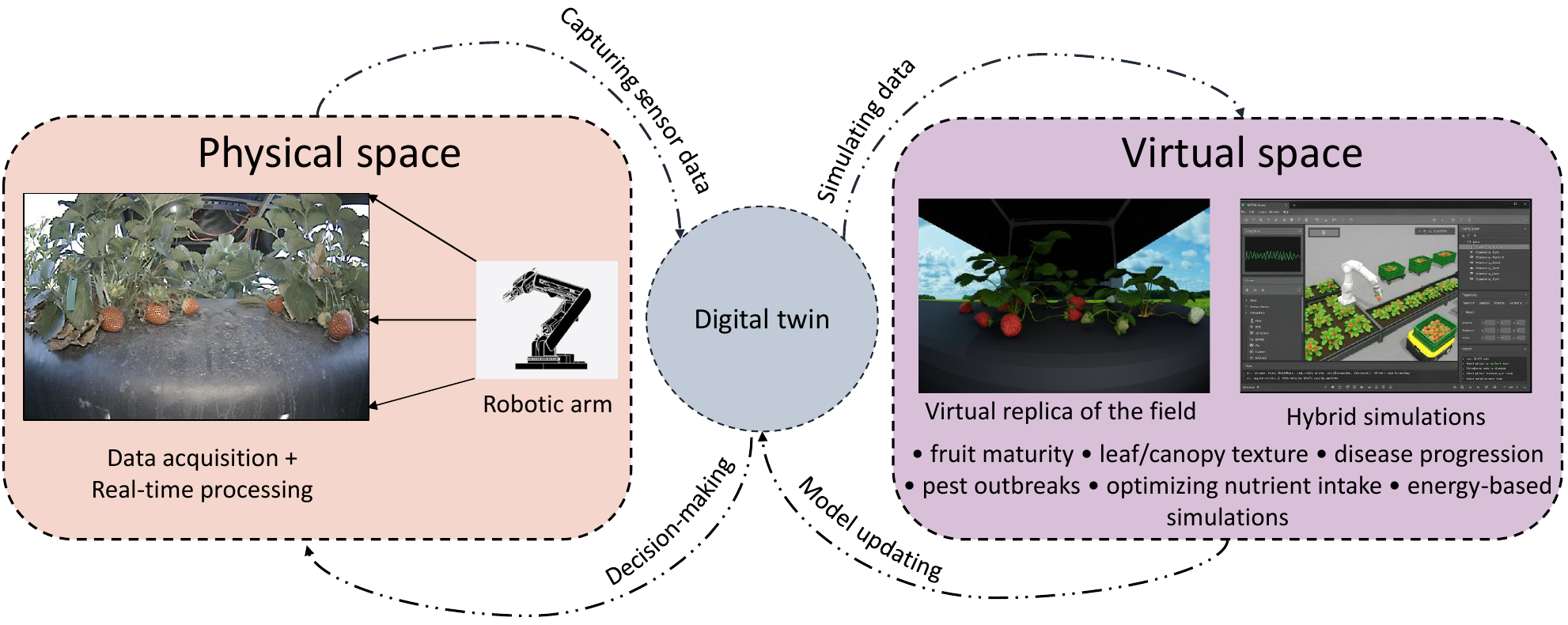}
    \caption{Schematic representation of a digital twin framework acting as a bridge between the physical and virtual spaces.~Real-world data from sensors are continuously synchronized with virtual simulations, enabling predictive analytics, scenario testing, and decision-making.~Insights from the digital (virtual) space are then transferred back to the physical system through actuator control, creating a closed-loop cycle of monitoring, simulation, and intervention.}
    \label{digitaltwin}
\end{figure}

Digital twin technology is rapidly gaining traction in precision agriculture as a means to model and simulate crop systems for better monitoring and decision-making.~A digital twin is essentially a virtual representation of a physical entity (crop or farm) that is continuously updated with real-world data~\cite{Mulhollem2024,mirbod2025simulation,kim2024agricultural,rajeswari2024digital,moghadam2020digital} (Fig.~\ref{digitaltwin}).~In agriculture, digital twins can integrate sensor readings, such as soil moisture, weather, crop health images with crop models to mirror the field's state in real time.~This allows growers and researchers to visualize conditions, run ``what-if'' scenarios, and predict future outcomes without risking the actual crop.~For example,~\cite{moghadam2020digital} describe a ``Digital-Twin Orchard'' concept: a virtual model of each tree in an orchard, paired with real-time data on that tree's condition. In their study, 3D LiDAR and camera systems on tractors were deployed to scan thousands of orchard trees and create their virtual counterparts in real time.~Over 15,000 mango, macadamia, avocado, and grapevine trees were digitized, and the data was used to model canopy characteristics and health indicators. Such a system enables continuous monitoring of orchard health, predicting plant stress, disease onset, and yield losses across the farm. 

While this project focused on monitoring (e.g. foliage density, light penetration) rather than automated spraying, it lays the groundwork for tree-specific disease management and AI-driven decision support platform, where one could test, for instance, how a new spraying regimen or pruning strategy might impact disease spread before applying it on the real orchard.~Several recent studies highlight the effectiveness of DTs for crop monitoring and simulation. In a 2024 study,~\cite{kim2024agricultural} built an agricultural digital twin for mandarin orange orchards. By aggregating big data from 185,000 hectares of mandarin farms into a virtual platform, they could analyze variations in fruit quality at regional, orchard, and even tree level. Their intra-orchard digital twin analysis explained fruit quality variation much better than broad inter-orchard data, demonstrating how digital twins enable ``micro-precision'' agriculture where each plant could receive customized treatment based on its digital profile.~Another example is the digital twin system for pepper pest management reported by~\cite{dai2024digital}.~This system created a virtual greenhouse of pepper plants infested with aphids, continuously fed by IoT sensors and camera data. The twin employed a predictive model (genetic algorithm-optimized random forest) to forecast pest population trends, and it evaluated control actions in simulation to recommend optimal intervention. In trials, the digital twin accurately predicted aphid population changes ($\approx$85\% accuracy) and helped optimize spraying decisions, improving economic outcomes by over 20\% compared to the growers' standard practice.

Despite these promising developments, the adoption of DTs in agriculture is still at an early stage. Reviews of the field indicate that many digital twin applications remain proof-of-concept or confined to lab/pilot studies. For instance, using digital twins for crop growth and disease predictive monitoring is largely in the research phase, and certain areas such as soil health management have seen only limited exploration~\cite{wang2024digital}.~Challenges include the complexity of biological systems, the need for extensive data integration, and the real-time synchronization required between the physical farm and its virtual counterpart.~Nevertheless, the trajectory is clear: as sensing technologies, data infrastructure, and modeling techniques advance, digital twins are poised to become integral to precision agriculture. They provide a holistic platform to monitor crop health, simulate interventions (like targeted spraying or irrigation changes), and foresee issues such as disease outbreaks before they happen.

\subsubsection{What are the benefits of combining reinforcement learning with digital twins in disease management?} \label{4.4.3}

The literature shows that RL algorithms can drive robotic sprayers to make smarter decisions pertaining to what, when, and how to treat, thereby improving precision and efficiency in the field. At the same time, digital twins supply high-fidelity virtual environments that mirror farm conditions, which is exactly what an RL agent needs to learn effective policies safely and efficiently~\cite{goldenits2024current}. Instead of deploying untested algorithms on real crops (with potentially costly errors), one can train an RL-based sprayer agent within a digital twin simulation of the crop, disease spread, and sprayer dynamics. The RL agent can interact millions of times with the twin (trying different spray timings, dosages, paths, etc.) to learn an optimal disease-control strategy, while the twin's models ensure the scenarios remain realistic. Researchers have started experimenting with such integrations. For example,~\cite{goldenits2024current} note that this synergy between environment simulation and learning is paving the way for ``reinforcement learning-based digital twin'' applications in farming.~In addition,~\cite{luo2025adaptive} demonstrated a digital twin-driven vertical farming system in which a Q-learning algorithm optimizes production scheduling inside the twin.~By linking the RL model with the virtual farm which was continuously updated with real sensor data, they achieved up to 78\% demand fulfillment in lettuce production, outperforming static optimization methods.~This study, while focused on yield and resource use, illustrates the general approach of embedding RL in a digital twin to handle dynamic decision problems. In the context of crop protection and spraying,~\cite{khosravi2025optimizing} built a simulated crop field infected with a pathogen and trained a hierarchical RL agent to manage a robotic sprayer within this digital environment. 

The RL agent had two levels of decision-making, a high level that decided where the robot should go next in the field, and a lower level that fine-tuned the path and spraying action on the detected diseased spots. The agent learned a policy that maximized disease coverage while minimizing wasted chemicals and energy. When benchmarked, the RL-driven approach significantly outperformed a conventional uniform spraying strategy with higher crop yield recovery with less pesticide across various infection scenarios and sensor noise levels. Furthermore, the twin can be used to continually retrain or fine-tune the RL policy as new data comes in, creating an adaptive system that improves with time. In practice, an autonomous sprayer could have its onboard AI ``living'' partly in the cloud-based digital twin: it tests various spray plans in the twin using the latest field state and then executes the best plan in the field, receiving real-world feedback to update the twin. This kind of RL-twin integration could handle disease outbreaks in a proactive manner (e.g., predicting where a fungal disease will spread next and preemptively guiding the robot to that area, or experimenting with different bio-control measures virtually before applying them). The joint use of RL and digital twins is expected to tackle many agricultural challenges and enable more efficient, adaptive farming processes, ultimately moving crop protection from reactive to predictive and optimized~\cite{goldenits2024current}.

A clear future direction is the development of fully autonomous crop protection systems that can detect, decide, and act in a closed loop. In such a system, high-resolution crop imaging (from drones or field cameras) would feed into a digital twin that continuously updates the location and severity of diseases or pests. An RL agent, possibly trained through thousands of simulated outbreak scenarios in the twin, would then determine the optimal intervention – whether that's a targeted spray, releasing a bio-control organism, adjusting irrigation to reduce pathogen spread, or some combination. This decision would be executed by robots or smart machinery in the field, and the outcomes (e.g. disease reduction, any side effects) would be measured and fed back into the twin for the next cycle. Over time, this self-learning approach could handle new diseases or evolving pest resistance by exploring adaptive strategies in simulation before applying them on the farm. The result would be a disease forecasting and proactive spraying system that is preventative rather than reactive and improves with experience, much like how a human crop scout gains intuition over many seasons, except here the intuition is encoded in AI and augmented by data from an entire network of farms.

\section{Challenges and Opportunities of Foundation Models for Precision Crop Disease Management}

Although FMs are offering multi-modalities of data formats to be processed, it still struggles with a few challenges within crop disease and pest management domain. These are associated in-field challenges that FMs right now cannot address or is incompetent to achieve any forms of success. 

\begin{enumerate}
    \item[a.] \textbf{Crop disease and pest data requires verification by an expert pathologist:} While FMs such as Stable Diffusion or GPT-based image generators can produce large numbers of high-resolution synthetic images of crop diseases, the quality of these images heavily depends on the accuracy of the training data.~Verification by an expert pathologist remains critical to ensure that the synthetic images accurately reflect real disease symptoms.~Moreover, in-field identification is inherently challenging due to environmental variability, overlapping symptoms, and early-stage subtle manifestations. In many cases, lab-based confirmation, such as PCR testing or pathogen isolation, is necessary to validate the diagnosis.
    
    \item[b.] \textbf{Confusion between visually similar disease symptoms:} Many text-to-image generative FMs may struggle to differentiate between visually similar disease symptoms, potentially producing unrealistic or misleading images. For example, in watermelons, early anthracnose lesions may resemble damage from leaf miners. Similarly, in tomatoes, early bacterial spot infections can appear similar to early blight lesions, and in wheat, Septoria leaf blotch can be mistaken for tan spot. Such confusions underscore the need for careful curation and verification of training datasets to ensure the FM learns accurate symptom representations.
    %\item[c] \textbf{Failing to impart realism achieved through in-field data collection:} While FMs can generate realistic looking disease images, it cannot impart.
    %\item[d] \textbf{Large reasoning models for autonomous operations may struggle in dynamic field conditions:} Since agriculture is by far one of the most dynamic areas where AI could be applied, diseases in specific is more dynamic and progressing as time changes.
    
    \item [c.] \textbf{Simulation fidelity and biological complexity:}~Digital twins promise to mirror crop and disease dynamics in real time, but their effectiveness hinges on the fidelity of underlying models.~Biological systems are inherently variable, and integrating multi-modal sensor data into accurate, continuously updated simulations remains a challenge.~Reflecting the stochastic nature of pest outbreaks, weather effects, and host-pathogen interactions in virtual environments remains a critical hurdle.
    
    \item [d.] \textbf{Safe and efficient reinforcement learning:}~Applying RL in real-world agriculture is constrained by safety and efficiency concerns. Training an RL agent directly in the field risks crop losses, excess chemical use, or equipment damage. While digital twins provide a safer training ground, the design of appropriate reward functions, coverage of diverse scenarios, and transfer of policies from simulation to field (the sim-to-real gap) remain open research problems.
    
    \item [e.] \textbf{Toward integrated closed-loop systems:}~Most current research still treats perception, simulation, and decision-making separately. Achieving robust closed-loop disease management requires unifying FMs for perception, DTs for simulation, and RL for adaptive decision-making in dynamic environments. The complexity of building such cyber-physical systems, resilient to weather variability, pathogen evolution, and incomplete data, is both a challenge and an opportunity for the field.
    
    \item [f.] \textbf{Opportunities ahead:} Despite these barriers, opportunities are significant.~Hybrid pipelines that combine FM-based perception, DT-based forecasting, and RL-based decision-making could transform pest and disease management from reactive spraying to proactive intervention. Regional or multi-farm digital twin networks could facilitate collaborative forecasting of outbreaks, while adaptive RL policies could evolve alongside shifting disease pressures and resistance patterns.~Integrating these frameworks with other site-specific management domains, such as irrigation and bio-control, could ultimately yield autonomous, resilient systems for sustainable crop protection.
\end{enumerate}

\section{Conclusion}

In conclusion, FMs are reshaping SSDM in crops by addressing key limitations of traditional machine and deep learning methods. They enable effective analysis of multi-modal datasets, connect textual descriptions to visual symptoms, and support interactive decision- making tools for farmers and extension personnel. The integration of adaptive and imitation learning in robotic systems further opens the door to precise, autonomous, and field-ready autonomous systems. Recent literature shows rapid growth in VLM research, highlighting their increasing importance. Although reinforcement learning and adaptive learning applications are still in early stages, combining them with digital twin simulations offers promising opportunities for testing and optimizing targeted strategies. Overall, these developments indicate that FMs will play a critical role in advancing intelligent, data-driven, and practical solutions for in-field crop monitoring and management.  

\section*{CRediT authorship statement}

\textbf{Nitin Rai:}~Data curation, Formal analysis, Investigation, Methodology, Writing-original draft.~\textbf{Daeun (Dana) Choi:}~Conceptualization, Writing-review \& editing.~\textbf{Nathan S. Boyd:}~Methodology, Writing-review \& editing.~\textbf{Arnold W. Schumann:}~Supervision, Funding acquisition, Writing-review \& editing.

\section*{Declaration of competing interest}

The authors declare that they have no known competing financial interests or personal relationships that could have appeared to influence the work reported in this paper. 

\section*{Acknowledgment}

This research was supported by the United States Department of Agriculture (USDA)-Small Business Innovation Research \& Technology Transfer Programs (SBIR/STTR) grant \# 2024-51402-42007.~Additionally, the authors declare that generative AI (ChatGPT, OpenAI, San Francisco, CA, USA) was used to improve grammar and language clarity during the preparation of this manuscript.

\bibliographystyle{unsrt}  
\bibliography{references}

\end{document}